\documentclass{article} 
\usepackage{iclr2025_conference,times}

\usepackage{amsmath,amsfonts,bm}

\def\eqref#1{equation~\ref{#1}}

\def\1{\bm{1}}

\DeclareMathAlphabet{\mathsfit}{\encodingdefault}{\sfdefault}{m}{sl}
\SetMathAlphabet{\mathsfit}{bold}{\encodingdefault}{\sfdefault}{bx}{n}

\usepackage{hyperref}
\usepackage{graphicx}
\usepackage{url}
\usepackage{booktabs}
\usepackage{caption}

\iclrfinalcopy

\title{IPDreamer: Appearance-Controllable 3D \\ Object Generation with Complex Image Prompts}

\author{
\textbf{Bohan Zeng{$^{1}$}\thanks{These authors contributed equally.}, \ Shanglin Li{$^1$}\footnotemark[1], \ Yutang Feng{$^1$}\footnotemark[1], \ Ling Yang{$^{5  \dag}$}, \ Hong Li{$^{1}$}} \\ \textbf{Sicheng Gao{$^{1}$}, \ Jiaming Liu{$^{3}$}, \ Conghui He{$^{7}$}, \ Wentao Zhang{$^{5}$}, \ Jianzhuang Liu{$^2$}}\\ \textbf{Baochang Zhang{$^{1,4}$}\thanks{Corresponding Author: bczhang@buaa.edu.cn, yangling0818@163.com.}, \ Shuicheng Yan{$^{6}$}} \\
    {$^1$}Institute of Artificial Intelligence, Beihang University \\ {$^2$}Shenzhen Institute of Advanced Technology, Shenzhen, China \\ {$^{3}$}Tiamat AI \ {$^{4}$}Zhongguancun Laboratory, Beijing, China \\ {$^{5}$}Peking University \ {$^{6}$}Skywork AI {$^{7}$}Shanghai Artificial Intelligence Laboratory \\
\url{https://github.com/zengbohan0217/IPDreamer}
}

\begin{document}

\maketitle

\begin{abstract}
  Recent advances in 3D generation have been remarkable, with methods such as DreamFusion leveraging large-scale text-to-image diffusion-based models to guide 3D object generation. These methods enable the synthesis of detailed and photorealistic textured objects. However, the appearance of 3D objects produced by such text-to-3D models is often unpredictable, and it is hard for single-image-to-3D methods to deal with images lacking a clear subject, complicating the generation of appearance-controllable 3D objects from complex images. To address these challenges, we present IPDreamer, a novel method that captures intricate appearance features from complex \textbf{I}mage \textbf{P}rompts and aligns the synthesized 3D object with these extracted features, enabling high-fidelity, appearance-controllable 3D object generation. Our experiments demonstrate that IPDreamer consistently generates high-quality 3D objects that align with both the textual and complex image prompts, highlighting its promising capability in appearance-controlled, complex 3D object generation. 

\end{abstract}

\section{Introduction}
\label{sec:intro}

The rapid evolution of 3D technology has revolutionized the way we create and interact with virtual worlds. 3D technology is now essential in a wide range of fields, including architecture, gaming, mechanical manufacturing, and AR/VR. However, creating high-quality 3D content remains a challenging and time-consuming task, even for experts. To address this challenge, researchers have developed text-to-3D methodologies that automate the process of generating 3D assets from textual descriptions. Built on the 3D scene representation capabilities of Neural Radiance Fields (NeRFs) \citep{mildenhall2021nerf, muller2022instant} and the rich visual prior knowledge of pretrained diffusion models \citep{rombach2022high, saharia2022photorealistic}, recent research \citep{jain2022zero, mohammad2022clip, poole2022dreamfusion, lin2023magic3d, chen2023fantasia3d, wang2023prolificdreamer, shi2023mvdream} has made significant progress, simplifying the text-to-3D pipeline and making it more accessible, which causes a significant shift in these fields.

Recent advances in diffusion models have significantly enhanced the capabilities of text-to-image generation. State-of-the-art (SOTA) systems, leveraging cutting-edge diffusion-based techniques \citep{nichol2021glide, rombach2022high, brooks2023instructpix2pix, zhang2023adding, hu2021lora}, can now generate and modify images directly from textual descriptions with vastly improved quality. Inspired by the rapid development in text-to-image generation, recent works \citep{poole2022dreamfusion, lin2023magic3d, chen2023fantasia3d, wang2023prolificdreamer} have extended these models to 3D by utilizing pretrained text-to-image diffusion models in conjunction with the Score Distillation Sampling (SDS) algorithm and its variants to optimize 3D representations. These methods are capable of generating high-quality 3D objects and scenes. However, due to the lack of explicit appearance information in textual prompts, the appearance of the generated results remains largely uncontrollable, limiting the precision of the visual output.

\begin{figure*}
  \centering
    \includegraphics[width=0.98\linewidth]{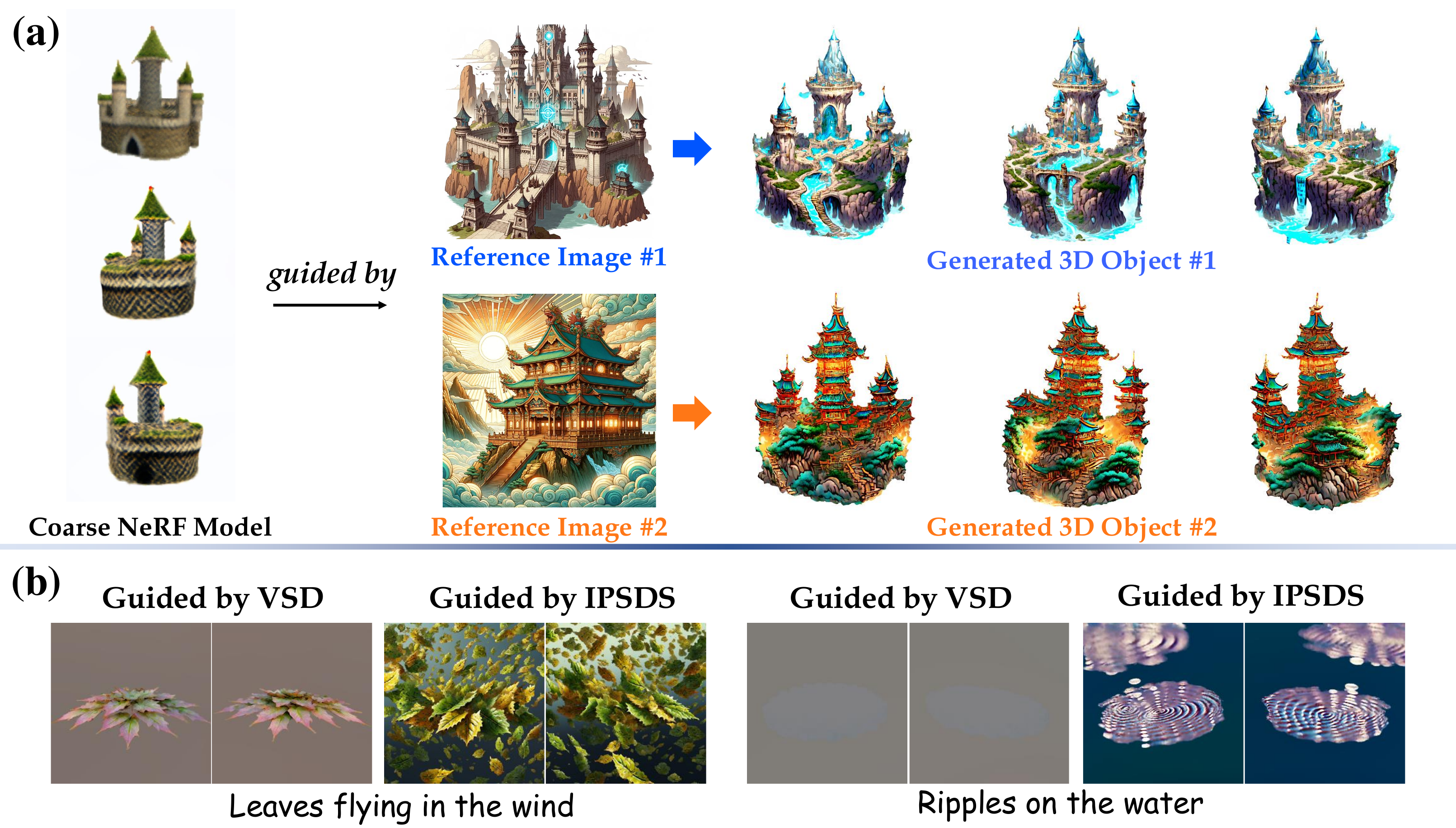}
    \caption{IPDreamer can generate controllable, high-quality 3D objects based on both textual and image prompts. (a) illustrates two high-quality 3D objects with rich details, initialized by the same NeRF model and guided by different complex reference image prompts. (b) demonstrates the 3D synthesis under challenging textual conditions, where our method outperforms existing text-to-3D method \citep{wang2023prolificdreamer}.
    }
    \label{fig:first}
    \vspace{-4mm}
\end{figure*}

Unlike the unpredictability in text-to-3D generation, single-image-to-3D generation allows for strict control over the appearance of the generated 3D results. However, existing single-image-to-3D methods \citep{liu2023zero, liu2023syncdreamer, shi2023mvdream} are limited to simple images with clear subjects, often struggling with complex images that feature rich content and intricate compositions. For example, they struggle with the complex images shown in Fig.\ref{fig:first}(a). Furthermore, as demonstrated in Fig.\ref{fig:first}(b), when text prompts are ambiguous or lack a clear main subject—such as “Leaves flying in the wind”—neither current text-to-3D nor image-to-3D methods can achieve reasonable 3D synthesis.

To tackle these challenges, we introduce \textit{IPDreamer}, a novel method for complex image-to-3D generation. Specifically, we extend SDS to \textit{Image Prompt Score Distillation Sampling (IPSDS)}, which leverages detailed features extracted from complex image prompts and corresponding normal maps to guide the optimization of both 3D mesh texture and geometry. With IPSDS, IPDreamer enables high-quality 3D object generation with controllable appearances based on complex image inputs. To ensure stable 3D object generation across various challenging scenarios, we propose a mask-guided compositional alignment strategy for IPSDS, enabling 3D object generation from multiple complex image prompts. In particular, we leverage a Multimodal Large Language Model (MLLM) to localize the features from multiple image prompts onto the generated 3D objects. This allows IPDreamer to handle diverse situations, including cases where multiple highly divergent images are used to guide the synthesis of a single 3D object, and scenarios where the guiding complex image exhibits significant structural differences from the initial coarse 3D object. As shown in Fig.\ref{fig:first}(a), IPDreamer effectively transfers the appearances of reference images to NeRF models, generating high-quality 3D objects even in cases with ambiguities or unclear primary subjects. For the more challenging scenarios illustrated in Fig.\ref{fig:first}(b), IPDreamer successfully produces the desired 3D objects where existing text-to-3D and single-image-to-3D methods fall short.

In summary, the main contributions of this paper are as follows:

\begin{itemize}

    \item We present IPDreamer, a novel 3D object synthesis framework that allows users to consistently create controllable, high-quality 3D objects. Compared with previous methods, IPDreamer excels in synthesizing high-quality 3D objects that closely align with complex image prompts.

    \item We introduce Image Prompt Score Distillation Sampling (IPSDS), which utilizes a substantial image prompt feature to guide 3D mesh optimization. 

    \item We propose a Mask-guided Compositional Alignment strategy for IPSDS, enabling high-quality 3D objects synthesis based on multiple complex image prompts, in cases where initial NeRF models deviate significantly from the provided image prompts, or when multiple diverse image prompts are needed to guide the synthesis of a single 3D object.

    \item Comprehensive experiments show that IPDreamer achieves high-quality 3D generation and excellent rendering results, outperforming existing SOTA methods.
    
\end{itemize}

\section{Related Work}
\label{sec:rel}

\subsection{Diffusion Models}

Diffusion models (DMs) were initially introduced as a generative model for gradually denoising images corrupted by Gaussian noise to generate samples \citep{sohl2015deep}. Recent advancements in DMs \citep{ho2020denoising, song2020denoising, dhariwal2021diffusion, vahdat2021score, rombach2022high, peebles2022scalable} have shown their exceptional performance in image synthesis. DMs have also achieved state-of-the-art results in various synthesis tasks, including text-to-image generation \citep{saharia2022photorealistic, nichol2021glide, ramesh2022hierarchical, podell2024sdxl, yang2024mastering}, inpainting \citep{avrahami2022blended, lugmayr2022repaint, ye2023ip}, 3D object synthesis \citep{li2022diffusion, luo2021diffusion}, video synthesis \citep{ho2022video, ho2022imagen}, speech synthesis \citep{kong2020diffwave, liu2021diffsinger}, super-resolution \citep{li2022srdiff, saharia2022image, gao2023implicit}, face animation \citep{qi2023difftalker}, text-to-motion generation \citep{tevet2022human}, and brain signal visualization \citep{takagi2022high, takagi2023high}. Some DMs \citep{kulikov2022sinddm, wang2022sindiffusion} can produce diverse results by learning the internal patch distribution from a single image. \citep{mokady2023null, tumanyan2023plug, wu2023tune, geyer2023tokenflow} enhance image/video editing with pre-trained DMs in a zero-shot or one-shot manner. These advancements highlight the versatility and potential of DMs across a wide range of syntheses.

\subsection{Controllable Generation and Editing}

Controllable generation and editing of 2D images and 3D objects are core goals of generative tasks. With the emergence of large language models (LLMs) such as GPT-3 and Llama \citep{brown2020language, touvron2023llama, touvron2023llama2}, instruction-based user-friendly generative control has gained much attention. InstructPix2Pix \citep{brooks2023instructpix2pix} and MagicBrush \citep{zhang2023magicbrush} build datasets based on LLMs and large text-to-image models to achieve effective instruction control on 2D images. InstructNeRF2NeRF \citep{haque2023instruct} combines this method with NeRF scene reconstruction \citep{mildenhall2021nerf} to introduce instruction control into 3D generation. 
Meanwhile, a series of adapter methods such as ControlNet and IP-Adapter \citep{zhang2023adding, hu2021lora, mou2023t2i, zhao2023uni, ye2023ip,huang2023composer} provide reliable approaches for fine-tuning large pre-trained DMs (e.g., Stable Diffusion \citep{rombach2022high} and Imagen \citep{saharia2022photorealistic}) for conditional controllable generation (e.g., using sketch, canny, pose, etc. to control image structure). Among them, image prompt adaption methods \citep{ye2023ip, zhang2023ssr} introduce a decoupled cross-attention mechanism to achieve effective appearance generation control using image prompts.

\subsection{3D Generation}
In recent years, 3D generative modeling has attracted a large number of researchers. Inspired by the recent neural volume rendering, many 3D-aware image rendering methods \citep{chan2022efficient, chan2021pi, gu2021stylenerf, hao2021gancraft, nguyen2019hologan, niemeyer2021giraffe} are proposed to generate high-quality rendered 2D images for 3D visualization. Meanwhile, with the development of text-to-image synthesis, researchers have shown a growing interest in text-to-3D generation. Early methods such as DreamField \citep{jain2022zero} and CLIPmesh \citep{mohammad2022clip} achieve text-to-3D generation by utilizing a pretrained image-text aligned model CLIP \citep{radford2021learning}. They optimize the underlying 3D representations (NeRFs and meshes) to ensure that all 2D renderings have high text-image alignment scores.
Recently, \citep{poole2022dreamfusion, lin2023magic3d, chen2023fantasia3d, wang2023prolificdreamer, chen2023text2tex} have achieved high-quality 3D synthesis (NeRFs and meshes) by leveraging a robust pretrained text-to-image DM as a strong prior to guiding the training of the 3D model. Other works \citep{shi2023mvdream, zhao2023efficientdreamer, liu2023zero} introduce multi-view DMs to enhance 3D consistency and provide strong structured semantic priors for 3D synthesis. IT3D \citep{chen2023it3d} combines SDS and GAN to refine the 3D model and obtain high-quality 3D synthesis. \citep{tang2023dreamgaussian, liang2023luciddreamer} combine 3D Gaussians \citep{kerbl20233d} with SDS-based optimization to improve 3D synthesis and reduce generation time. Additionally, \citep{melas2023realfusion, tang2023make, liu2023one, qian2023magic123} are capable to generate 3D representations based on single images, and \citep{liu2023zero, liu2023syncdreamer, shi2023mvdream, yang2023consistnet} achieve 2D images in multiple viewpoints, which enable consistant 3D object generation.
In this work, we introduce IPDreamer, a method that leverages complex image prompts to provide comprehensive appearance information, effectively guiding the synthesis of high-quality 3D objects.
\vspace{-2mm}

\begin{figure*}[tb]
  \centering
    \includegraphics[width=0.96\linewidth]{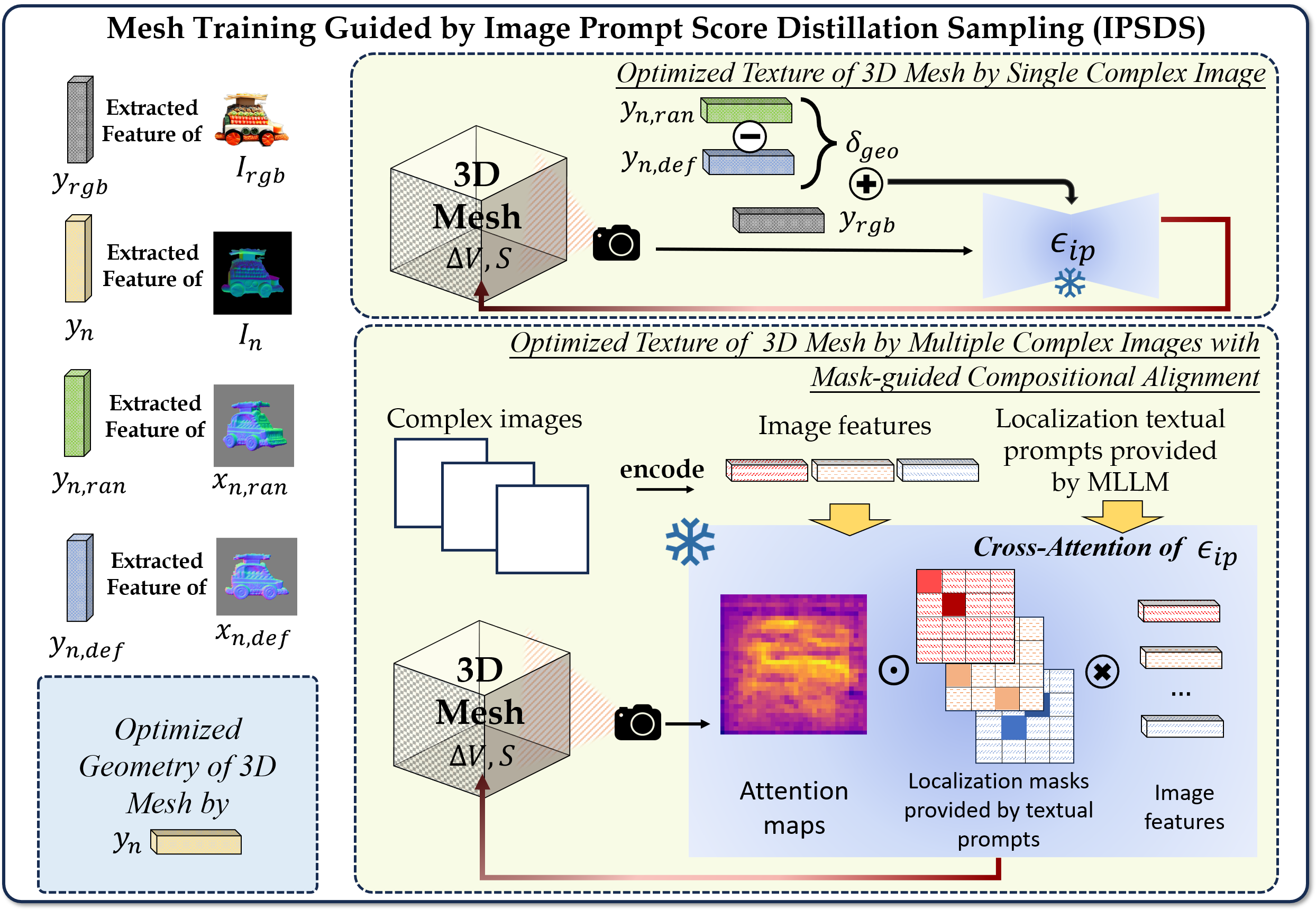}
    \caption{
    IPDreamer is designed to generate high-quality, appearance-controllable 3D meshes that align with single/multiple complex image prompts.}
    \label{fig:overview}
    \vspace{-4mm}
\end{figure*}

\section{Method}
\label{sec:method}
\vspace{-2mm}

In this section, we present the details of IPDreamer. We start with a brief definition of 3D mesh, followed by the problem statement for text-to-3D and single-image-to-3D generation, and a review of SDS preliminaries. Next, we present the design and analysis of our proposed IPSDS and Mask-guided Compositional Alignment. Note that 3D meshes are the most common form of 3D representation in industry, so our approach focuses on optimizing 3D meshes.
\vspace{-2mm}

\subsection{Preliminaries and Motivation}

\paragraph{3D Mesh.}
The 3D mesh can be represented as a deformable tetrahedral grid $(V, T)$, where each vertex $v_i \in V$ has a signed distance field value $s_i \in S$ and a deformation $\Delta v_i \in \Delta V$ from its canonical position. During optimization, the surface mesh is rendered into high-resolution images using a differentiable rasterizer \citep{munkberg2022extracting}.
\vspace{-2mm}

\paragraph{Score Distillation Sampling.} 
Given a textual prompt $y$ or an image $I$, text-to-3D/single-image-to-3D generation aims to synthesize novel views and optimize the parameters of a 3D object/scene corresponding to the given $y$ or $I$.
DreamFusion \citep{poole2022dreamfusion} utilizes a pretrained text-to-image DM $\epsilon_{pretrain}$ to optimize an MLP parameterized as $\theta$ representing a 3D volume, where a differentiable generator $g$ renders $\theta$ to create 2D images $x=g(\theta, c)$ given a sampled camera pose $c$, based on the gradient of the Score Distillation Sampling (SDS) loss:
\begin{equation}
    \nabla_\theta \mathcal{L}_{\mathrm{SDS}}(\theta)=\mathbb{E}_{t, \epsilon}\left[w(t)\left(\epsilon_{pretrain} \left(x_t ; y, t\right)-\epsilon\right) \frac{\partial x}{\partial \theta}\right],
  \label{eq:SDS}
\end{equation}
\noindent where $w(t)$ is a weighting function, $\epsilon_{pretrain} \left(x_t ; y, t\right)$ predicts the noise $\epsilon \sim \mathcal{N}(\rm{0}, \rm{I})$, given the noisy image $x_t$, text prompt features $y$ and timestep $t$.

\paragraph{Motivation.} Although \citep{lin2023magic3d, chen2023fantasia3d, wang2023prolificdreamer} show excellent text-to-3D generation, the appearances of their 3D synthesis results are uncontrollable. 
A feasible solution to realize controllable 3D object generation is to use a 2D image as a prior. However, existing single-image-to-3D methods find obtaining high-quality 3D object synthesis from complex images difficult. To overcome these obstacles, we extract feature details from complex image prompts to provide comprehensive appearance information for 3D object synthesis, and we propose the IPSDS and a Mask-guided Compositional Alignment strategy to enable stable, high-fidelity 3D object generation.


\subsection{Image Prompt Score Distillation Sampling (IPSDS)}

In this section, we leverage the image features extracted by the image encoder $\mathcal{E}_{image}$ from \citep{ye2023ip} to optimize the geometry and texture of 3D objects. The 3D mesh is initialized using either a user-provided 3D mesh, or a coarse NeRF model generated by existing text-to-3D methods or IPSDS. As shown in Fig.~\ref{fig:overview}, we first explain how IPSDS optimizes $\Delta V$, $S$, and $\theta$. Subsequently, we analyze how IPSDS effectively utilizes complex image prompts to guide the synthesis of high-quality 3D objects.

\paragraph{Optimizing 3D Mesh with IPSDS.}
Existing methods directly use a text-conditioned DM to guide geometry optimization. However, it can be challenging because the DM's pre-training dataset lacks normal map images. To address this, we adopt an additional normal image prompt feature $y_{n}=\mathcal{E}_{image}(I_{n})$ to provide richer and more robust geometric information with normal map optimization, instead of solely using the textual prompt $y$ \citep{chen2023fantasia3d, wang2023prolificdreamer}. 


The geometry optimization process computes the gradients of the IPSDS geometry loss as:
\begin{equation}
\begin{aligned}
    &\nabla_{\Delta V} \mathcal{L}_{\mathrm{IPSDS}-Geo}(\Delta V, S)= 
    \mathbb{E}_{t, \epsilon}[w(t) \ (\epsilon_{ip} (z_{n,t} ; y_{n}, y, t)-\epsilon) \frac{\partial z_{n}}{\partial \Delta V}], \\
    \label{eq:stage2_geo_1}
\end{aligned}
\vspace{-5mm}
\end{equation}
\begin{equation}
\begin{aligned}
    &\nabla_{S} \mathcal{L}_{\mathrm{IPSDS}-Geo}(\Delta V, S)= 
    \mathbb{E}_{t, \epsilon}[w(t) \ (\epsilon_{ip} (z_{n,t} ; y_{n}, y, t)-\epsilon) \frac{\partial z_{n}}{\partial S}],
  \label{eq:stage2_geo_2}
\end{aligned}
\end{equation}

where $z_{n,t}$ denotes the noisy latent of rendered normal map in random view $x_{n,ran}$ at timestep $t$. 

After optimizing the estimated normal map, the geometry of the 3D mesh becomes more reasonable. 
Then we further optimize the texture through IPSDS.
We first extract the image prompt features $y_{rgb}=\mathcal{E}_{image}(I_{rgb})$, as a basic guidance for the texture optimization. 
Then we devise a geometry prompt difference $\delta_{geo}$ for $y_{rgb}$ to compensate for the morphological disparity between $x_{rgb}$ and $I_{rgb}$.
Let $x_{n, def}$, and $x_{n, ran}$ be the rendered normal map of the 3D mesh from the default viewpoint and a randomly sampled viewpoint, respectively.
We extract their image prompt featuress, ${y}_{n,def}=\mathcal{E}_{image}(x_{n, def})$ and ${y}_{n,ran}=\mathcal{E}_{image}(x_{n,ran})$. The difference between ${y}_{n,ran}$ and ${y}_{n,def}$ is called the geometry prompt difference $\delta_{geo}$:
\begin{equation}
\begin{aligned}
    \delta_{geo} = y_{n,ran} - y_{n,def}.
  \label{eq:stage2_geo_delta}
\end{aligned}
\end{equation}

The texture optimization process computes the gradients of the IPSDS texture loss as:
\begin{align}
\begin{split}
    &\nabla_{\theta} \mathcal{L}_{\mathrm{IPSDS}-Tex}(\theta, \Delta V, S)=  
    \mathbb{E}_{t, \epsilon}[w(t) \ (\epsilon_{ip} (z_{rgb,t} ; y_{rgb}+\delta_{geo}, y, t)-\epsilon) \frac{\partial z_{rgb}}{\partial \theta}],
  \label{eq:stage2_tex_1}
\end{split}  \\
\begin{split}
    &\nabla_{\Delta V} \mathcal{L}_{\mathrm{IPSDS}-Tex}(\theta, \Delta V, S)=  
    \mathbb{E}_{t, \epsilon}[w(t) \ (\epsilon_{ip} (z_{rgb,t} ; y_{rgb}+\delta_{geo}, y, t)-\epsilon) \frac{\partial z_{rgb}}{\partial \Delta V}],
  \label{eq:stage2_tex_2}
\end{split} \\
\begin{split}
    &\nabla_{S} \mathcal{L}_{\mathrm{IPSDS}-Tex}(\theta, \Delta V, S)=  
    \mathbb{E}_{t, \epsilon}[w(t) \ (\epsilon_{ip} (z_{rgb,t} ; y_{rgb}+\delta_{geo}, y, t)-\epsilon) \frac{\partial z_{rgb}}{\partial S}],
  \label{eq:stage2_tex_3}
\end{split}
\end{align}

where $z_{rgb,t}$ denotes the noisy latent of $x_{rgb}$ (random viewpoint) in timestep $t$. The geometry prompt difference $\delta_{geo}$ can effectively represent the Morphological distance between $x_{n,ran}$ and $x_{n,def}$ in the image prompt feature space. Thus it is used to compensate $y_{rgb}$ (default viewpoint) such that $y_{rgb}+\delta_{geo}$ represents the RGB image $x_{rgb}$.

\paragraph{Incorporating Image Prompt into 3D Generation.}
Here we explain how our method can effectively use a complex, high-quality image to guide 3D object synthesis, by introducing cross-attention for the image prompt. Given the query features $Z$ which are derived from the latent representations of the 2D rendering results of the 3D object from various viewpoints, and the image prompt embedding $y_{rgb}$, the cross-attention for the image prompt is formulated as follows:
\begin{equation}
\begin{aligned}
    Z' = \mathrm{Softmax}(\frac{\mathbf{Q}\mathbf{K}^{\top}}{\sqrt{d}})\mathbf{V},
\end{aligned}
\end{equation}
where $\mathbf{Q}=Z \mathbf{W}_{q}$, $\mathbf{K} = y_{rgb} \mathbf{W}^{ip}_{k}$, $\mathbf{V} = y_{rgb} \mathbf{W}^{ip}_{v}$ represent the queries, keys, and values within the cross-attention module, respectively, $Z'$ denotes the output features of the module, and the $\mathbf{W}_{q}$, $\mathbf{W}^{ip}_{k}$, and $\mathbf{W}^{ip}_{v}$ are the projection matrices used for linear transformations. The reasons why IPSDS can utilize complex images to guide the generation of 3D objects while existing single-image-to-3D methods cannot are two-fold: First, the encoder of the image prompt adaption method effectively extracts the image features $y_{rgb}$ from the reference high-resolution image prompt. Secondly, as the attention map can accurately align the features $y_{rgb}$ with specific positions of rendered images \citep{hertz2022prompt} from the 3D object, the features from the original complex image are precisely positioned on the most relevant parts of the 3D object.



\begin{figure}
    \centering
    \includegraphics[width=0.95\linewidth]{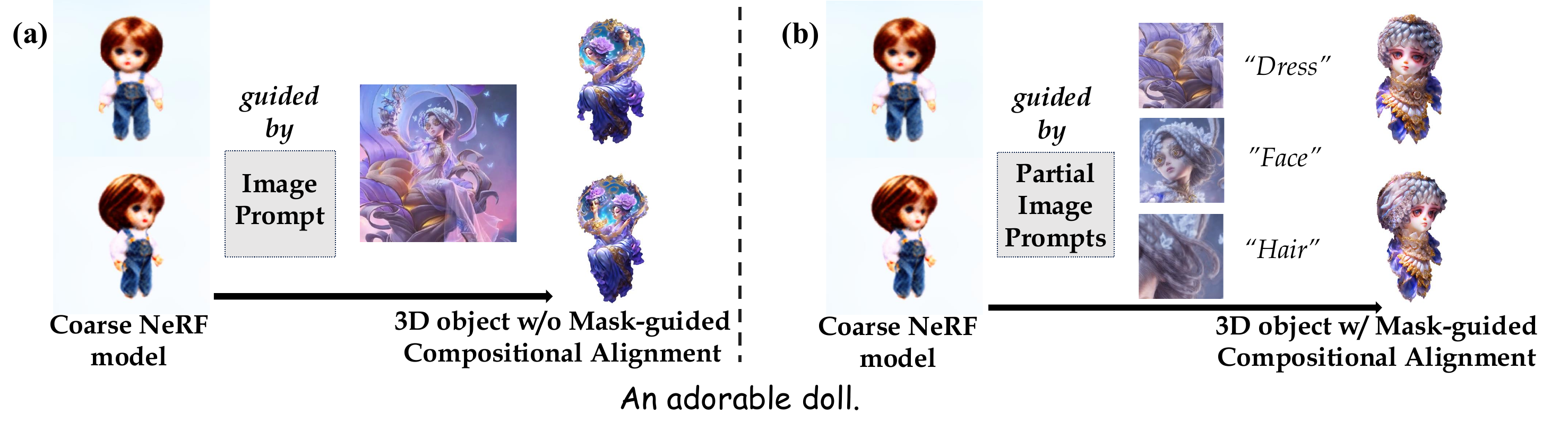}
    \caption{Illustration of the effectiveness of Mask-guided Compositional Alignment.}
    \label{fig:CIPA}
    \vspace{-4mm}
\end{figure}

\subsection{Mask-guided Compositional Alignment for Multiple Image Prompts}

\paragraph{Motivation of Mask-guided Compositional Alignment.} When the rendered 2D images from the NeRF model significantly differ from the image prompt or when the input includes multiple diverse image prompts, relying solely on cross-attention for the image prompt fails to effectively align the features $y_{rgb}$ with specific positions on the 3D object. As illustrated in Fig.~\ref{fig:CIPA}(a), it is evident that IPDreamer encounters difficulties with this challenging sample. To address this issue, we have designed a Mask-guided Compositional Alignment strategy. Specifically, we collect multiple images $I^{rgb}_{i}$ from the input complex image prompts. Then, we employ a large multimodel model (GPT-4v) to provide localization words $y^{txt}_{i}$ for corresponding $I^{rgb}_{i}$. We adopt the cross attention in LDM~\citep{rombach2022high} to obtain localization masks:
\begin{equation}
\begin{aligned}
  &m_{i} = \mathrm{BI}(\mathrm{Softmax}(\frac{\mathbf{Q}\mathbf{K}^{\top}_{txt,i}}{\sqrt{d}})), \
  \mathbf{Q} = Z \mathbf{W}_{q}, \ \mathbf{K}_{txt,i}=y^{txt}_{i} \mathbf{W}^{txt}_{k},  \ i=1,2,...,n_{ip}, 
  \label{eq:sc_txt_mask}
\end{aligned}
\end{equation}
where $\mathrm{BI}$ denotes a binarization operator and $n_{ip}$ is the number of the input multiple images.
Subsequently, the mask $m_i$, obtained from the textual description $y^{txt}_{i}$, is used to adjust the computation of the cross attention corresponding to the feature $y^{rgb}_{i}$ of the image prompt $I^{rgb}_{i}$:
\begin{equation}
\begin{aligned}
  &Z' = \frac{1}{n_{ip}} \sum_{i=1}^{n_{ip}} m_{i} \ \mathrm{Softmax}(\frac{\mathbf{Q}\mathbf{K}^{\top}_{ip,i}}{\sqrt{d}}) \mathbf{V}_{ip,i},
  \label{eq:sc_attn}
\end{aligned}
\end{equation}
where $\mathbf{Q}=Z \mathbf{W}_{q}$, $\mathbf{K}_{ip,i}=y^{rgb}_{i}\mathbf{W}^{ip}_{k}$, $\mathbf{V}_{ip,i}=y^{rgb}_{i} \mathbf{W}^{ip}_{v}$. With the help of our strategy, we successfully localize the features of the multiple images onto the 3D object, as shown in Fig.~\ref{fig:CIPA}(b). Next, we provide more details of the IPSDS training process with the Mask-guided Compositional Alignment.

\paragraph{Training Process of IPSDS with Mask-guided Compositional Alignment.}
We briefly describe how IPSDS is designed with the Mask-guided Compositional Alignment. First, we utilize GPT-4v to generate localization textual prompts, $y^{txt}_{1}, ..., y^{txt}_{n_{ip}}$, to map features of the complex images onto 3D objects. Specifically, we input the multiple complex reference images and rendered images of a coarse NeRF model into GPT-4v, which analyzes and identifies the regions that need to be segmented from the input complex images and generates the corresponding localization textual prompts. Based on the analysis, we employ SAM \citep{kirillov2023segment} to segment multiple partial images, $I^{rgb}_{1}, ..., I^{rgb}_{n_{ip}}$, from the complex images. Additionally, both the localization textual prompts, $y^{txt}_{1}, ..., y^{txt}_{n_{ip}}$, and the segmented partial images, $I^{rgb}_{1}, ..., I^{rgb}_{n_{ip}}$, can be adjusted by users.

Given potential semantic differences between $I_{rgb}$ and the coarse NeRF model (e.g., “magnificent magic castle” vs. “adorable cottage”) and the possibility that the multiple partial images $I^{rgb}{1}$,…,$I^{rgb}{n_{ip}}$ may lack detail or resolution, it is crucial to enhance them before initiating texture optimization.
We adopt a super-resolution model \citep{zhang2023adding} \footnote{\url{https://huggingface.co/lllyasviel/control_v11f1e_sd15_tile}} in conjunction with $I^{rgb}_{1},...,I^{rgb}_{n_{ip}}$ and $y^{txt}_{1}$,..., $y^{txt}_{n_{ip}}$ to generate new $I^{rgb}_{1},...,I^{rgb}_{n_{ip}}$. This preprocessing step improves the quality of both the guided images and the resulting 3D object.

\begin{figure}[tb]
    \centering
    \includegraphics[width=0.7\linewidth]{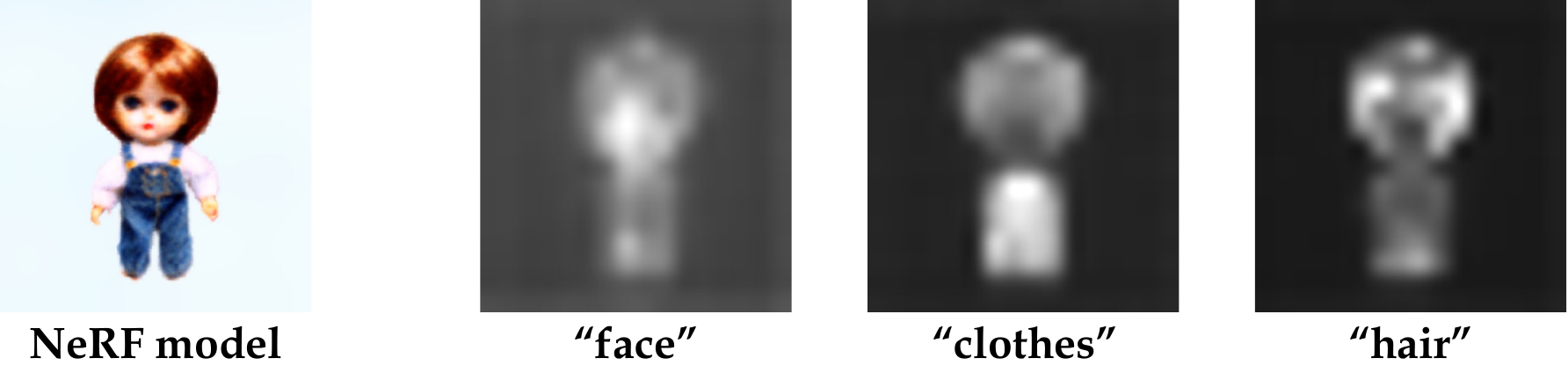}
    \caption{Visualization of localization masks.}
    \label{fig:attn_mask_sc}
    \vspace{-6mm}
\end{figure}

Subsequently, we extract image prompt features $y^{rgb}_{1},..., y^{rgb}_{n_{ip}}$ from corresponding partial image prompts $I^{rgb}_{1},..., I^{rgb}_{n_{ip}}$. Then, we localize the features onto the 3D object according to $y^{txt}_{1}$,..., $y^{txt}_{n_{ip}}$, based on Equation~\ref{eq:sc_txt_mask} and Equation~\ref{eq:sc_attn}. Fig.~\ref{fig:attn_mask_sc} shows an example of the effect of localization masks calculated in the process of Mask-guided Compositonal Alignment. The IPSDS supervision in this part can be written as:
{\small
\begin{align}
\begin{split}
&\nabla_{\theta} \mathcal{L}_{\mathrm{IPSDS}-Tex}(\theta, \Delta V, S)= 
\mathbb{E}_{t, \epsilon}[w(t) \ (\epsilon_{ip} (z_{rgb,t} ; y^{rgb}_{1},...,y^{rgb}_{n_{ip}}, y^{txt}_{ 1},...,y^{txt}_{n_{ip}}, t)-\epsilon) \frac{\partial z_{rgb}}{\partial \theta}], 
  \label{eq:stage2_sctex_1}
\end{split} \\
\begin{split}
&\nabla_{\Delta V} \mathcal{L}_{\mathrm{IPSDS}-Tex}(\theta, \Delta V, S)= 
\mathbb{E}_{t, \epsilon}[w(t) \ (\epsilon_{ip} (z_{rgb,t} ; y^{rgb}_{1},...,y^{rgb}_{n_{ip}}, y^{txt}_{ 1},...,y^{txt}_{n_{ip}}, t)-\epsilon) \frac{\partial z_{rgb}}{\partial \Delta V}], 
  \label{eq:stage2_sctex_2}
\end{split} \\
\begin{split}
&\nabla_{S} \mathcal{L}_{\mathrm{IPSDS}-Tex}(\theta, \Delta V, S)= 
\mathbb{E}_{t, \epsilon}[w(t) \ (\epsilon_{ip} (z_{rgb,t} ; y^{rgb}_{1},...,y^{rgb}_{n_{ip}}, y^{txt}_{ 1},...,y^{txt}_{n_{ip}}, t)-\epsilon) \frac{\partial z_{rgb}}{\partial S}]. 
  \label{eq:stage2_sctex_3}
\end{split}
\end{align}
}

After initially localizing the partial image prompts onto the 3D object, it is then necessary to further optimize the texture of the 3D object globally. We input all features of the partial images and the provided complex images into the IPSDS loss to optimize the 3D object simultaneously:
\begin{align}
\begin{split}
    f_{global} = \mathrm{concat}(y^{rgb}_{1},...,y^{rgb}_{n_{ip}}, y_{rgb}+\delta_{geo}), 
\end{split}\\
\begin{split}
    \nabla_{\theta} \mathcal{L}_{\mathrm{IPSD}-Tex}(\theta, \Delta V, S)=  
    \mathbb{E}_{t, \epsilon}[w(t) (\epsilon_{ip} (z_{rgb,t} ; f_{global}, t)-\epsilon) \frac{\partial z_{rgb}}{\partial \theta}], 
  \label{eq:stage2_scgtex_1}
\end{split}\\
\begin{split}
    \nabla_{\Delta V} \mathcal{L}_{\mathrm{IPSD}-Tex}(\theta, \Delta V, S)=  
    \mathbb{E}_{t, \epsilon}[w(t) (\epsilon_{ip} (z_{rgb,t} ; f_{global}, t)-\epsilon) \frac{\partial z_{rgb}}{\partial \Delta V}], 
  \label{eq:stage2_scgtex_2}
\end{split}\\
\begin{split}
    \nabla_{S} \mathcal{L}_{\mathrm{IPSD}-Tex}(\theta, \Delta V, S)=  
    \mathbb{E}_{t, \epsilon}[w(t) (\epsilon_{ip} (z_{rgb,t} ; f_{global}, t)-\epsilon) \frac{\partial z_{rgb}}{\partial S}]. 
  \label{eq:_stage2_scgtex_3}
\end{split}
\end{align}

\section{Experiments}
\label{sec:experi}

\begin{figure*}[tb]
  \centering
    \includegraphics[width=0.98\linewidth]{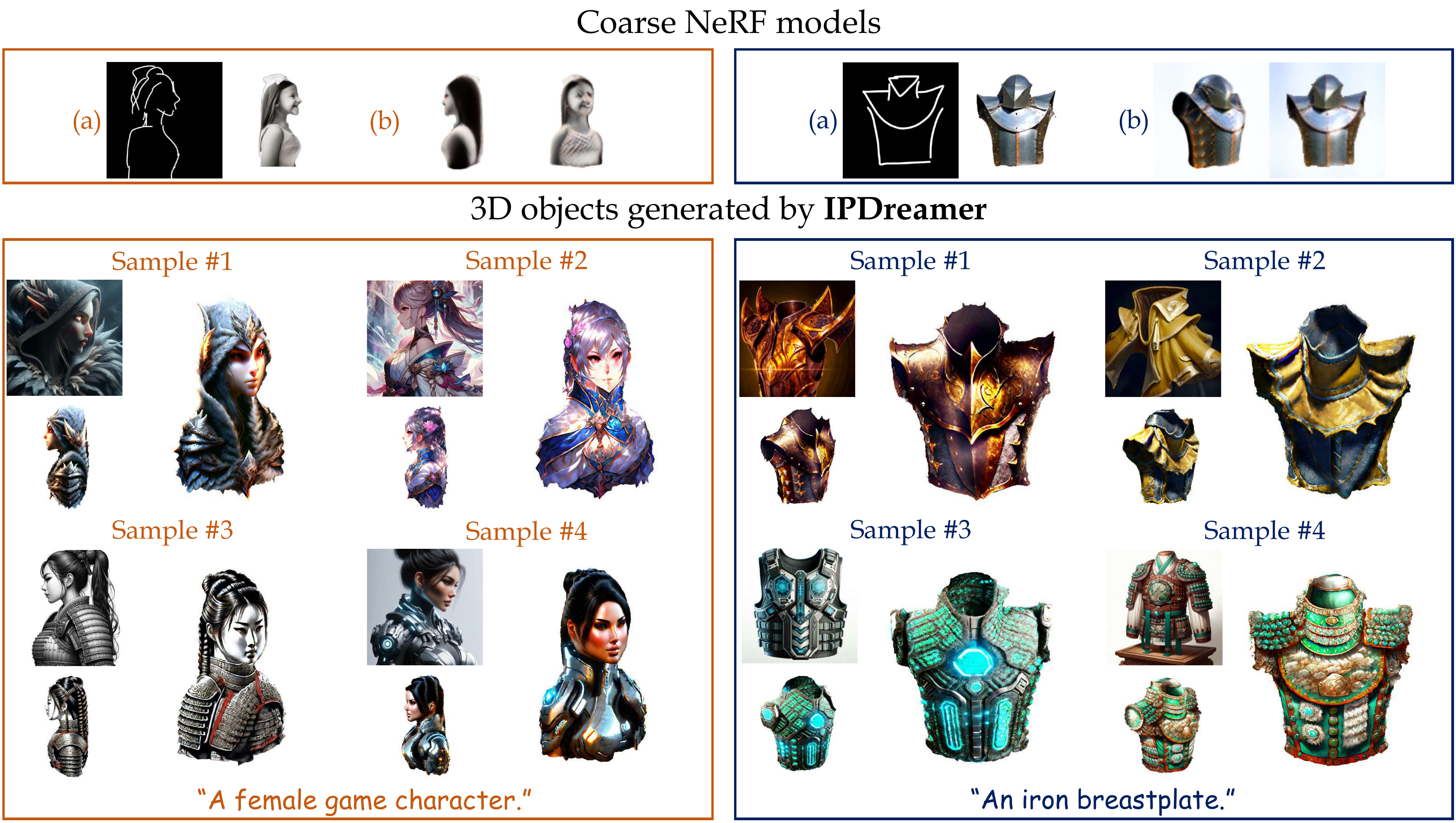}
    \caption{generated 3D objects with different image prompts. (a) Image prompts used for Coarse NeRF model generation. (b) Rendering of Coarse NeRF models. We show four samples for each textual prompt. In each sample, the top left is a selected complex image prompt, and the bottom left and the right illustrate the high-quality 3D object optimized by IPDreamer based on the coarse NeRF model.}
    \label{fig:texture_edit}
    \vspace{-3mm}
\end{figure*}

\subsection{3D Generation with Single Complex Image}

As depicted in Fig. \ref{fig:texture_edit}, we show generated 3D objects that use diverse image prompts to guide synthesis. This demonstrates IPDreamer's ability to produce high-quality 3D objects that align with the styles of the provided images. Remarkably, IPDreamer can appropriately transfer the appearance of the image prompts to the synthesized 3D objects, regardless of the structure difference between the image prompts and the coarse NeRF models. To our knowledge, this high-quality appearance transfer task is not achievable by existing text-to-3D or single-image-to-3D methods. 
In Sample 2 for the textual prompt ``An iron breastplate'', although both the textual and image prompt features are provided for 3D object synthesis, the generated result resembles a leather breastplate more closely, which aligns with the image prompt rather than the ``iron'' mentioned in the textual prompt. This illustrates that the image prompt exerts a stronger influence on the synthesis of the 3D object than the textual prompt. Consequently, such a powerful ability to edit 3D object textures greatly facilitates applications in the gaming and video industries.

\begin{figure*}[tb]
  \centering
    \includegraphics[width=0.98\linewidth]{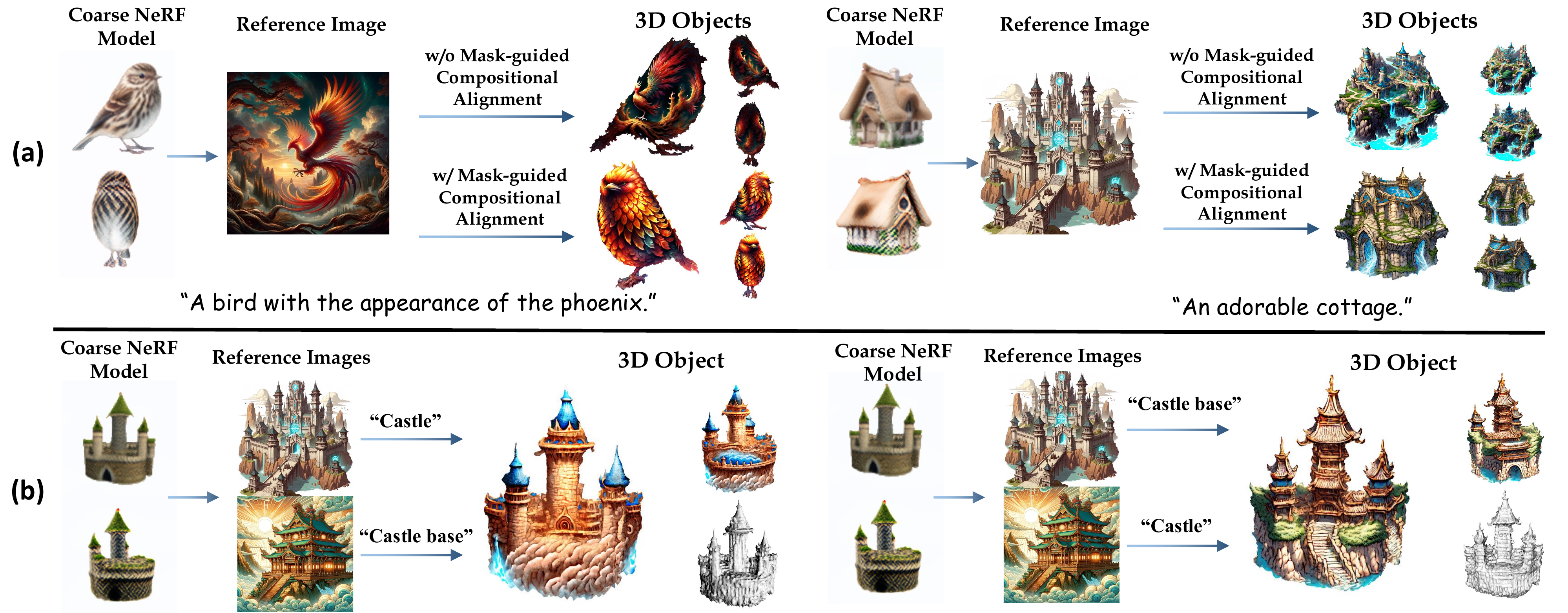}
    \caption{Effectiveness of Mask-guided Compositional Alignment.}
    \label{fig:mul_img_edit}
    \vspace{-3mm}
\end{figure*}


\begin{table}[h]

\begin{minipage}[tb]{0.45\linewidth}
    \centering
    \caption{Quantitive comparison of text-to-3D generation.}
    \label{tab:text23d_metric}
    \setlength{\tabcolsep}{1.5mm}\begin{tabular}{lccccc}
    \toprule
    Method          & FID $\downarrow$ & CLIP-Score $\uparrow$  \\ \hline
    DreamFusion      & 320.16 & 0.2413   \\
    Magic3D          & 320.56 & 0.2582   \\
    Fantasia3D       & 294.79 & 0.2557   \\
    ProlificDreamer  & 277.35 & 0.2603   \\
    LRM              & 304.81  & 0.2522   \\
    LGM              & 296.62  & 0.2605   \\
    IPDreamer(Ours)  & \textbf{253.32} & \textbf{0.2716}  \\ 
    \bottomrule
    \end{tabular}
\end{minipage}
\quad
\begin{minipage}[tb]{0.5\linewidth}
    \centering
    \caption{Percentage of the preference in the user study of text-to-3D generation.}
    \label{tab:user_study}
    \begin{tabular}{l|cc}
    \toprule
    Method           & Prefer baseline & Prefer ours \\ \hline
    DreamFusion      & 6.45   & \textbf{93.55}   \\
    Magic3D          & 10.89  & \textbf{89.11}   \\
    Fantasia3D       & 25.82  & \textbf{72.18}   \\
    ProlificDreamer  & 41.65  & \textbf{58.35}   \\
    LRM              & 29.73  & \textbf{70.27}   \\
    LGM              & 34.52  & \textbf{65.48}   \\
    \bottomrule
    \end{tabular}
\end{minipage}
\end{table}

\begin{figure*}[tb]
  \centering
    \includegraphics[width=0.98\linewidth]{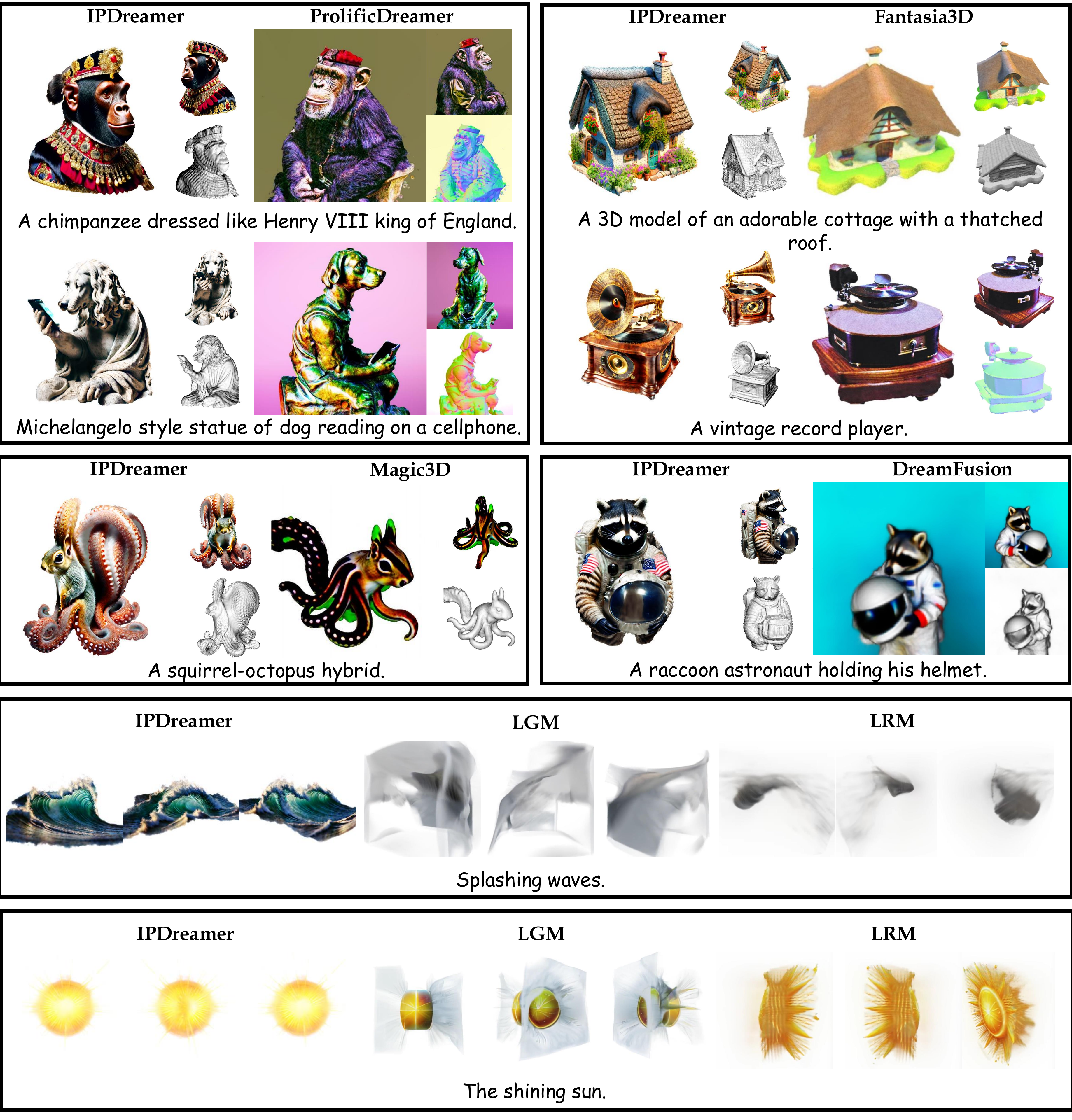}
    \caption{Qualitative Comparison of Text-to-3D Generation. It is worth noting that for the "Shining Sun" sample, our IPDreamer can generate a luminous 3D sphere with natural light rays emitting, which is difficult for other baseline methods to achieve.}
    \label{fig:text23d_comp}
    \vspace{-3mm}
\end{figure*}

\subsection{3D Generation with Multiple Complex Images}

To demonstrate the stability of our IPDreamer in generating 3D models guided by multiple complex images or when the initial coarse 3D object significantly differs from these guiding images, we produced more 3D objects under these conditions. As shown in Fig.~\ref{fig:mul_img_edit}(a), when provided image prompts vastly differ from the coarse NeRF models, the 3D objects guided by IPSDS with Mask-guided Compositional Alignment retain the semantic essence of the original coarse NeRF models and achieve the intended outcomes. And in Fig.~\ref{fig:mul_img_edit}(b), we provide two samples with the same coarse NeRF model and multiple diverse complex image prompts but different mask-guided textual prompts; the generated results of these two samples are quite different and well follow the textual requirements, showing that IPDreamer effectively enhances the diversity of the generated 3D objects, offering new perspectives for the advancement in the 3D research domain.
\vspace{-2mm}

\subsection{Comparison on Text-to-3D}
\label{subsec:text23d}

To validate the quality of the results generated by our method, we conducted a comparative analysis with baseline methods \citep{poole2022dreamfusion, lin2023magic3d, chen2023fantasia3d, wang2023prolificdreamer, hong2024lrm, tang2024lgm} in the text-to-3D generation task. As illustrated in Fig.~\ref{fig:text23d_comp}, IPDreamer surpasses these baseline methods by producing highly controllable and realistic 3D objects that align closely with the provided textual prompts. Additionally, we present examples such as ``The shining sun'' and ``Splashing waves'', where existing text-to-3D and single-image-to-3D methods fail to generate clear and coherent subjects. In contrast, our method successfully generates results that meet the requirements, further demonstrating its effectiveness. 

For a quantitative evaluation, we randomly select 30 textual prompts and compare the performance of IPDreamer against state-of-the-art (SOTA) methods, as shown in Table~\ref{tab:text23d_metric}. IPDreamer achieves superior performance, evidenced by a lower FID score, indicating higher quality 3D object generation, and a higher CLIP score, reflecting better alignment with the input textual prompts. To provide a more comprehensive assessment of the generated results, we also conduct a user study, the results are demonstrated in Table.~\ref{tab:user_study}. The details of the CLIP score, FID, and user study are introduced in Appendix~\ref{Sec:train_detail}.
\vspace{-1mm}

\subsection{Ablation Study}

\begin{figure}[t]
\centering
\noindent
\begin{minipage}[tb]{0.6\linewidth}
    \includegraphics[width=0.98\linewidth]{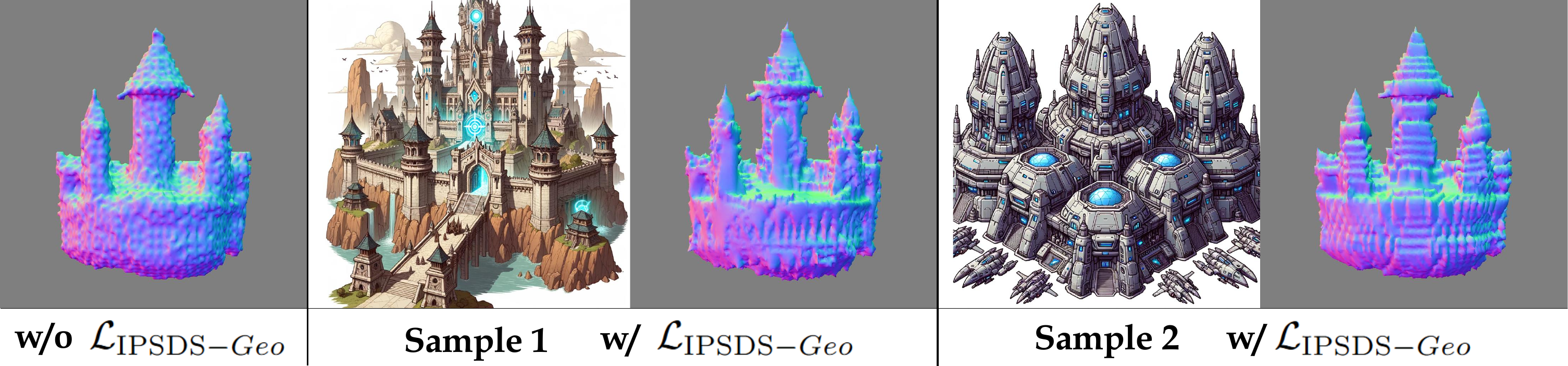}
    \caption{Visualization of the initial normal map of the 3D object at the beginning of geometry optimization, along with the image prompt and the refined normal map after geometry optimization for each sample.}
    \label{fig:ablation1}
\end{minipage}
\quad
\begin{minipage}[tb]{0.3\linewidth}
  \centering
    \captionof{table}{Ablation study of $\delta_{geo}$.}
    \label{tab:As_delta_geo}
    \setlength{\tabcolsep}{2mm}{
    \begin{tabular}{l|c}
    \toprule
     Methods                   & CLIP score $\uparrow$   \\ \hline
     w/o $\delta_{geo}$        & 0.8228    \\
     w/ $\delta_{geo}$          & \textbf{0.8389}   \\
    \bottomrule
    \end{tabular}}
\end{minipage}
\vspace{-3mm}
\end{figure}

We conduct an ablation study to evaluate the impact of $\mathcal{L}_{\mathrm{IPSDS}-Geo}$ and $\delta_{geo}$ on optimizing 3D objects. Their effectiveness is illustrated in Fig.~\ref{fig:ablation1} and Table~\ref{tab:As_delta_geo}.
In Fig.~\ref{fig:ablation1}, we showcase the optimized normal maps of two samples. After geometry optimization, Sample 1 and Sample 2 learn the high-frequency details from their corresponding image prompts. The difference in the optimized normal maps between Sample 1 and Sample 2 is readily discernible in Fig.~\ref{fig:ablation1}, illustrating the efficacy of $\mathcal{L}_{\mathrm{IPSDS}-Geo}$ in learning geometry representations from image prompts. 
In Table~\ref{tab:As_delta_geo}, we compare the CLIP score of 3D objects optimized with and without $\delta_{geo}$. We conduct the quantitive comparison using the samples mentioned in Section~\ref{subsec:text23d} and employ CLIP score to compare the alignment of rendered images of 3D objects generated with and without $\delta_{geo}$ in different viewpoints with the reference image prompt. The experimental results show that with $\delta_{geo}$, the rendered images of the 3D object in different viewpoints are more consistent with the reference image prompt. 
\vspace{-1mm}

\section{Conclusion}
\label{sec:conclu}

In this work, we propose IPDreamer, a novel framework that enables the generation of high-quality, appearance-controllable 3D objects from complex image prompts. By introducing Image Prompt Score Distillation Sampling (IPSDS), our method effectively captures rich and intricate appearance features from complex images to guide the optimization of both texture and geometry in 3D mesh generation. Our approach supports multiple complex images in various contexts to guide 3D object generation, enabling the stable production of high-quality 3D results. IPDreamer addresses the limitations of existing text-to-3D and single-image-to-3D methods by producing 3D objects that are consistent with textual descriptions and the detailed appearances of complex image prompts. Comprehensive experiments demonstrate that IPDreamer outperforms state-of-the-art methods, highlighting its promising capability in advancing appearance-controllable complex 3D object generation.

\bibliography{iclr2025_conference}
\bibliographystyle{iclr2025_conference}

\newpage

\appendix
\section{Appendix}

In Appendix~\ref{Sec:tex_edit}, we present additional synthesized 3D objects generated by IPDreamer. Detailed implementation information is provided in Appendix~\ref{Sec:train_detail}.  Furthermore, Appendix~\ref{Sec:impact} analyzes the social impact of IPDreamer.

\subsection{More Examples of 3D Objects Generated by IPDreamer}
\label{Sec:tex_edit}

\begin{figure}[h]
  \centering
    \includegraphics[width=\linewidth]{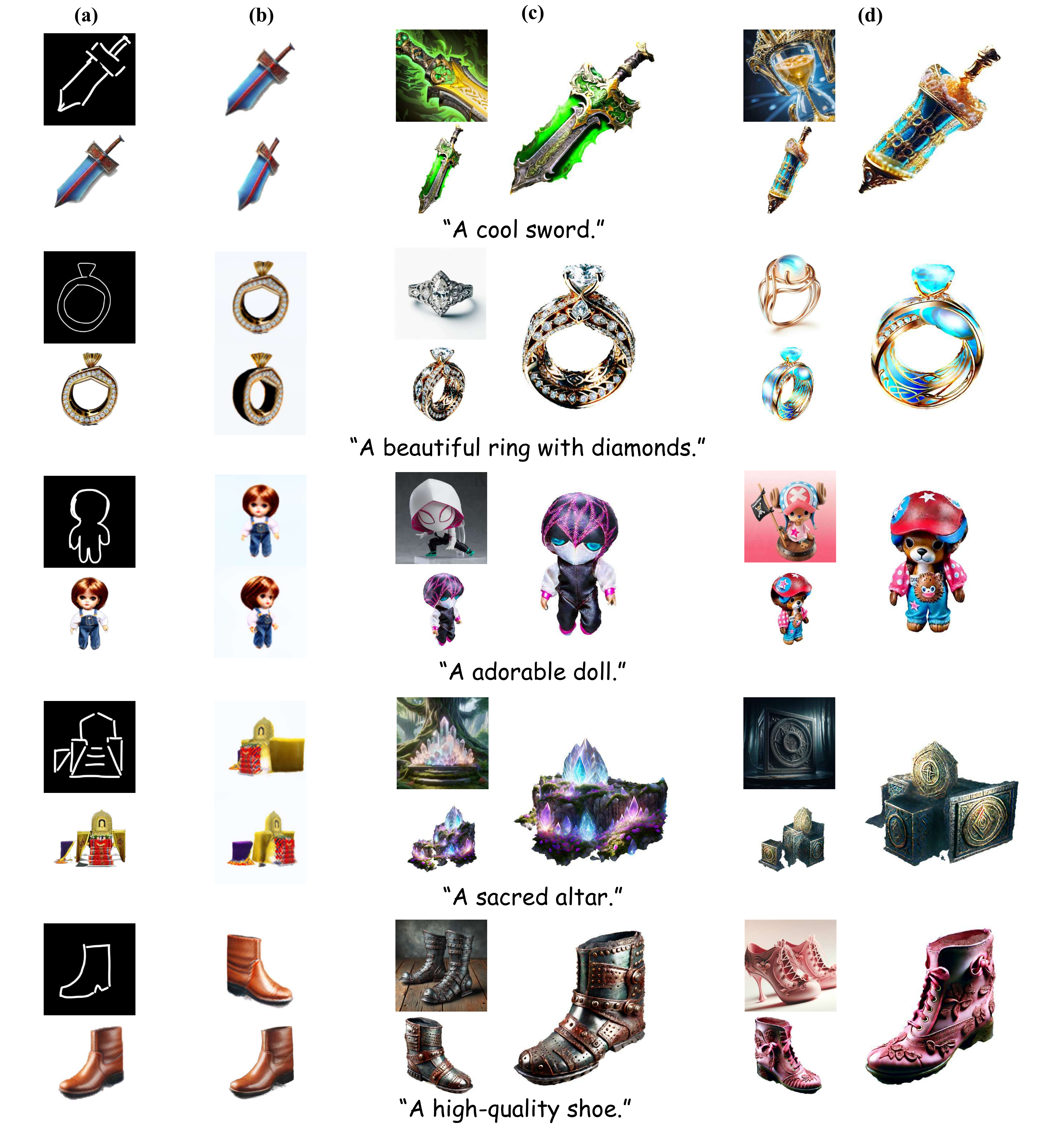}
    \caption{generated 3D objects with different image prompts. (a) Scribble object outlines and corresponding image prompts for Coarse 3D object generation. (b) Renderings of coarse NeRF models. (c) (d) Two samples demonstrated for each textual prompt. In each sample, the top left is a reference complex image prompt, and the bottom left and the right illustrate the 3D object optimized by IPDreamer based on the coarse NeRF model.}
    \label{fig:tex_edit_supp}
\end{figure}

\subsubsection{More Examples of 3D Objects Guided by IPSDS}

To further demonstrate IPDreamer's remarkable ability to manipulate appearance, we conduct more 3D object synthesis experiments. These experiments use diverse textual prompts, each accompanied by two distinct image prompts. As evident in Fig.~\ref{fig:tex_edit_supp}, IPDreamer consistently produces impressive 3D object synthesis, regardless of the geometric shape of the acquired NeRF or the image prompts used for texture editing. The generated results highlight IPDreamer's powerful texture editing capabilities for 3D objects, suggesting its potential to serve effectively in the 3D gaming and video industries.

\begin{figure}[tb]
    \centering
    \includegraphics[width=0.98\linewidth]{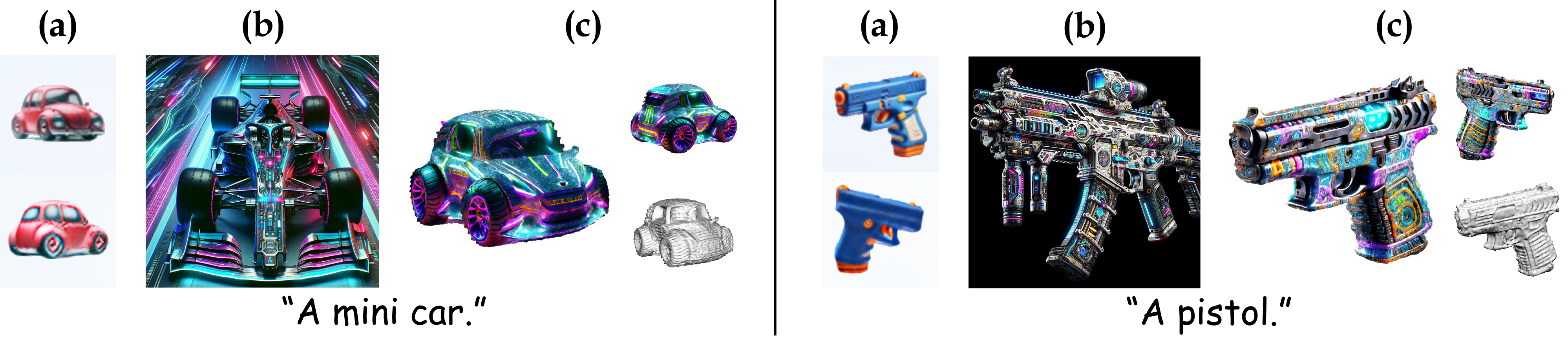}
    \caption{More samples of 3D object editing. (a) Coarse NeRF models. (b) Provided image prompts. (c) 3D objects generated by IPDreamer.}
    \label{fig:extram_texture_edit}
\end{figure}

\begin{figure}[tb]
    \centering
    \includegraphics[width=\linewidth]{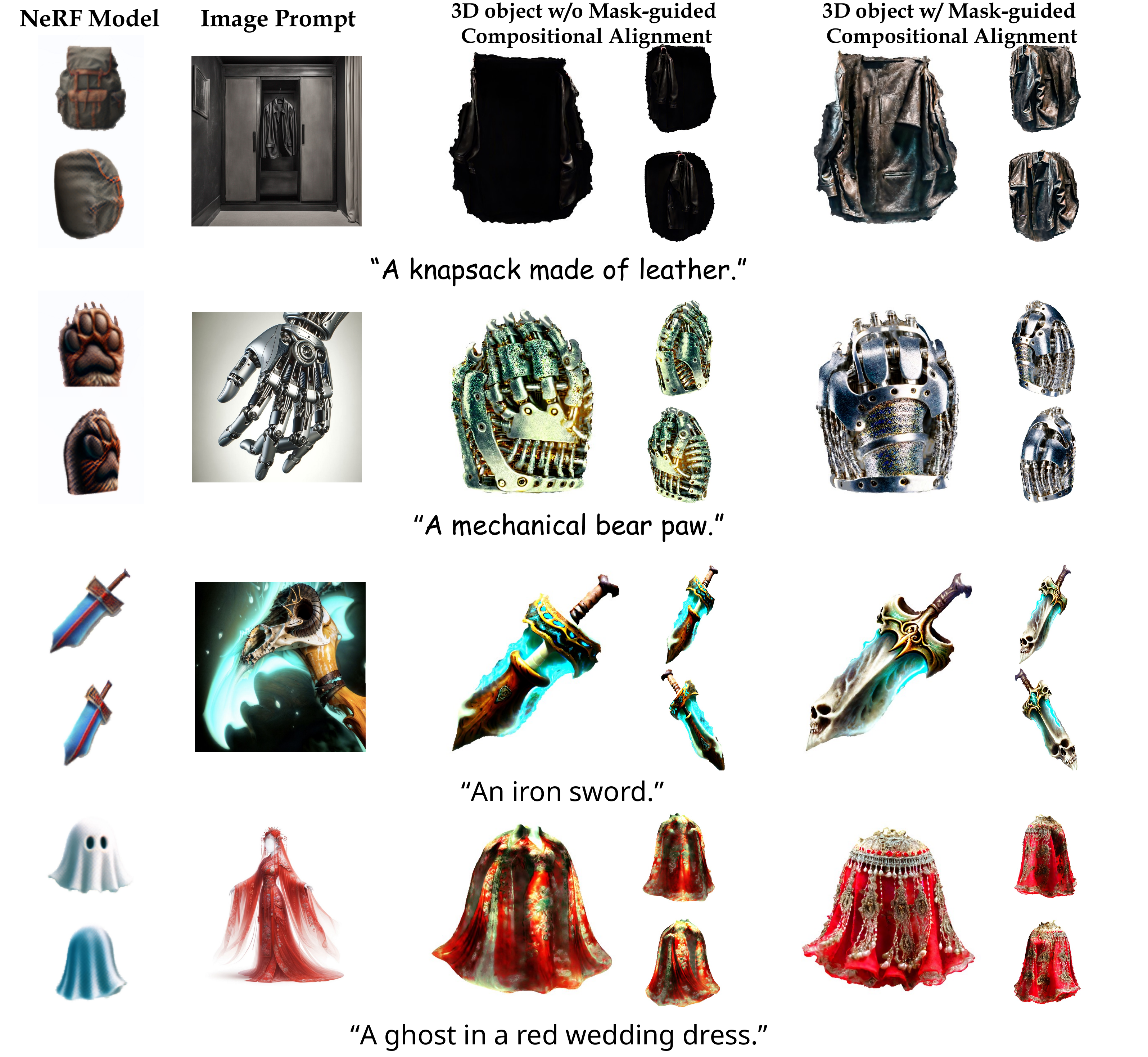}
    \caption{Comparison of 3D objects synthesis with and without Mask-guided Compositional Alignment.}
    \label{fig:supp_tex_edit_comp}
\end{figure}

\begin{figure}[tb]
    \centering
    \includegraphics[width=\linewidth]{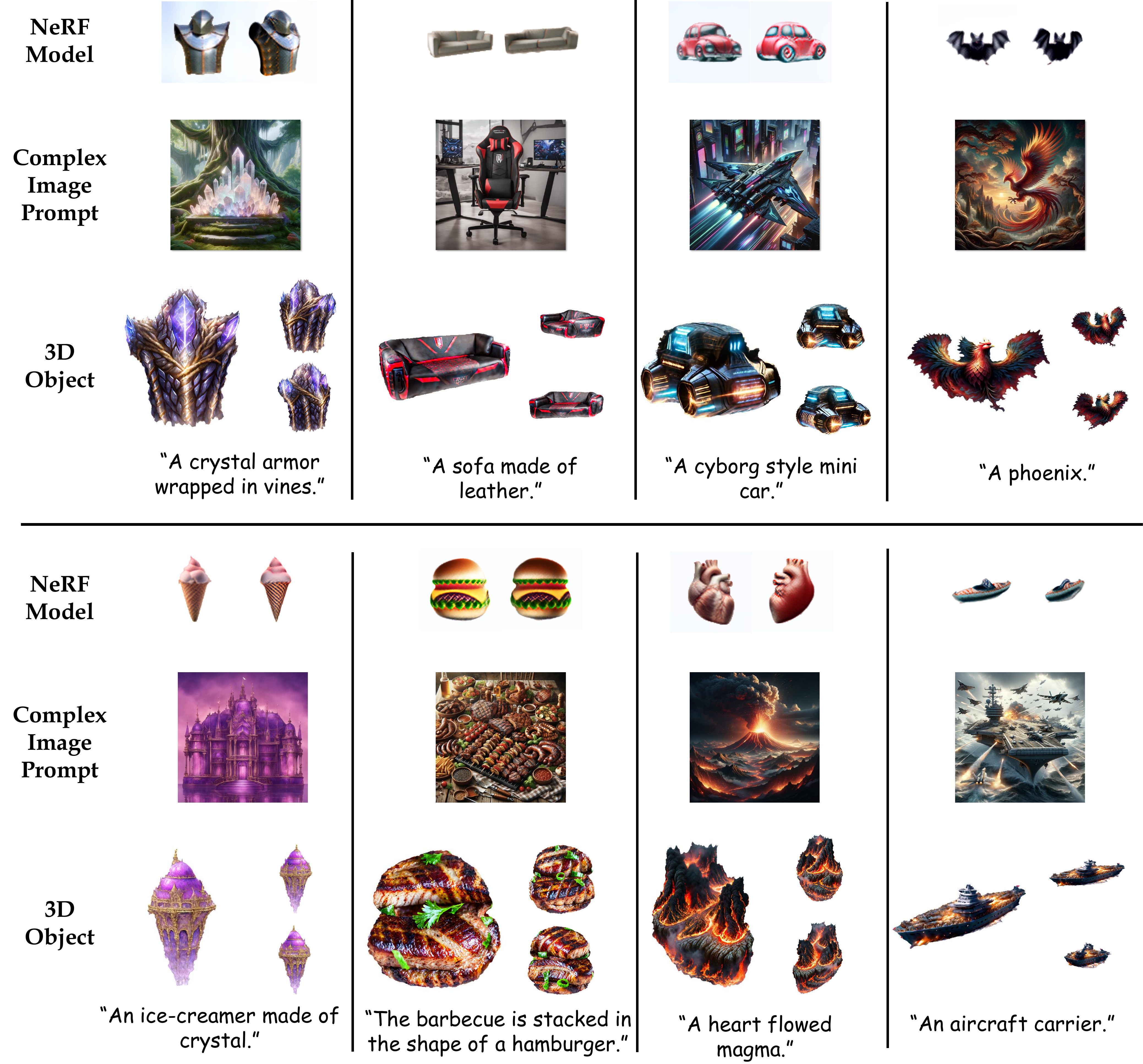}
    \caption{3D objects synthesis with Mask-guided Compositional Alignment. }
    \label{fig:tex_edit_1}
\end{figure}

Besides, to further demonstrate the appearance guidance capability of IPSDS in generating 3D objects, we use two samples whose reference image prompts are particularly complex and somewhat different from the initial coarse NeRF models, as shown in Fig.~\ref{fig:extram_texture_edit}. Even in such challenging cases, IPDreamer can still achieve high-quality 3D objects, such as the cyborg-style mini car generated in the first example, and the futuristic toy pistol in the second example. By utilizing IPDreamer's style editing ability for 3D objects, the generated results can be more diverse.

\subsubsection{More Examples of 3D Objects Guided with Mask-guided Compositional Alignment}

While IPDreamer can achieve remarkable 3D object synthesis in numerous challenging cases even without Mask-guided Compositional Alignment, difficulties emerge when the appearance of the supplied image prompts substantially diverge from the initial coarse NeRF model. To emphasize the potent 3D object optimization capability of Mask-guided Compositional Alignment within IPDreamer, we offer a comparison of the generated 3D objects with and without Mask-guided Compositional Alignment in Fig.~\ref{fig:supp_tex_edit_comp}. The outcomes validate the exceptional high-fidelity capability of Mask-guided Compositional Alignment. To further elucidate the superiority of Mask-guided Compositional Alignment, we present additional generation results in Fig.~\ref{fig:tex_edit_1}

\begin{table}[tb]
\centering
\resizebox{\textwidth}{!}{
\begin{tabular}{c}
\toprule
A praying mantis wearing roller. \\
Michelangelo-style statue of a dog reading news on a cellphone. \\
A matte painting of a castle made of cheesecake surrounded by a moat made of ice cream.  \\
A chimpanzee dressed like Henry VIII king of England.  \\
A 3D model of an adorable cottage with a thatched roof.  \\
A plate piled high with chocolate chip cookies.  \\
A vintage record player.  \\
A car made out of cheese.  \\
A beautifully carved wooden knight chess piece.  \\
A car made out of sushi.  \\
A squirrel-octopus hybrid.  \\
A small saguaro cactus is planted in a clay pot.  \\
A DSLR photo of an imperial state crown of England.  \\
A rotary telephone carved out of wood.  \\
A raccoon astronaut holding his helmet.  \\
A classic Packard car.  \\
A cauldron full of gold coins.  \\
A blue tulip.   \\
A stuffed grey rabbit holding a pretend carrot.  \\
A plush dragon toy. \\
A broken egg.       \\
A popped balloon.   \\
Leaves flying in the wind.     \\
A robot assembles itself.       \\
Lightning.        \\
The shining sun.      \\
A melting ice cube.  \\
Ripples on water.    \\
A broken bridge.     \\
Splashing waves.    \\
\bottomrule \\
\end{tabular}}
\caption{Textual prompts used in the quantitative comparison.}
\label{tab:used_prompt}
\end{table}

\subsection{Implementaion Details}
\label{Sec:train_detail}

\subsubsection{Optimization}

In this work, we conduct all of our experiments on one A100-SXM4-40GB GPU. In Stage~1, we optimize $5k$ steps with Adam optimizer \cite{xie2022adan} to obtain a NeRF model. In Stage~2, we optimize $10k$ steps for geometry optimization and $15k$ steps for texture optimization. During each optimization progress in Stage~2, we initially sample the timesteps $t \sim \mathcal{U}(0.02, 0.98)$ for the first $5k$ steps, and then sample $t$ from $t \sim \mathcal{U}(0.02, 0.5)$ for the rest steps. Each optimization process in Stage~2 requires approximately 9GB GPU memory with batch size 1 and a rendering resolution of 512.

\subsubsection{Textual Prompts used for Comparison}

We provide the 30 randomly selected textual prompts for quantitative comparison and user study in Table.~\ref{tab:used_prompt}. To fully compare the generation capabilities of different methods and demonstrate the effectiveness of our method, the testing textual prompts include 20 textual prompts that are frequently used in previous text-to-3D methods as well as 10 relatively challenging textual prompts that do not have a clear main subject.

\subsubsection{Metrics} 
We perform quantitative comparisons to evaluate IPDreamer's performance, with the following metrics: 
\begin{itemize}
    \item CLIP score \cite{gal2022stylegan}: We employ CLIP score in Section~4.2 of the main paper. By assessing the alignment between the textual descriptions and the rendered images of 3D objects from various viewpoints, we can judge whether text-to-3D methods successfully generate 3D objects that match the input textual prompts.

    \item Fréchet Inception Distance (FID) \cite{heusel2017gans}: To evaluate the quality of the generated results, we utilize FID to compare the similarity between the rendering images of 3D objects and the images generated by the text-to-image model, Stable Diffusion.
\end{itemize}

\subsubsection{User Study}

To further verify the quality of our generated results, we follow previous works \cite{lin2023magic3d, chen2023fantasia3d, wang2023prolificdreamer} and conduct a user study by comparing IPDreamer with the six SOTA methods \cite{poole2022dreamfusion, lin2023magic3d, chen2023fantasia3d, wang2023prolificdreamer, hong2024lrm, tang2024lgm}, under 16 prompts randomly selected from Table~\ref{tab:used_prompt}. 
Each of the 80 volunteers is provided with 16 pairs of results corresponding to the 16 prompts. In each pair, one from IPDreamer and one from a randomly selected baseline. Thus, there are a total of 1280 pairwise comparisons. The volunteers are then asked to choose the better result in terms of faithfulness, quality, and fidelity.

\subsection{Social Impact}
\label{Sec:impact}

Our IPDreamer does not have a direct negative impact on society. However, it is important to recognize the potential of high-quality 3D objects and ensure they are not adopted for malicious purposes.

\end{document}